\documentclass[sigconf]{acmart}

\AtBeginDocument{%
  \providecommand\BibTeX{{%
    \normalfont B\kern-0.5em{\scshape i\kern-0.25em b}\kern-0.8em\TeX}}}

\usepackage{tikz}
\newcommand*\circled[1]{\tikz[baseline=(char.base)]{
            \node[shape=circle,fill,inner sep=1pt] (char) {\footnotesize \textcolor{white}{#1}};}}

\usepackage[boxed,ruled]{algorithm2e}

\usepackage{mathtools}
\DeclarePairedDelimiter{\ceil}{\lceil}{\rceil}
\usepackage{multicol}
\usepackage{wrapfig}
\usepackage{subfig}
\copyrightyear{2022}
\acmYear{2022}
\setcopyright{acmcopyright}\acmConference[DAC '22]{Proceedings of the 59th
ACM/IEEE Design Automation Conference (DAC)}{July 10--14, 2022}{San Francisco,
CA, USA}
\acmBooktitle{Proceedings of the 59th ACM/IEEE Design Automation Conference
(DAC) (DAC '22), July 10--14, 2022, San Francisco, CA, USA}
\acmPrice{15.00}
\acmDOI{10.1145/3489517.3530585}
\acmISBN{978-1-4503-9142-9/22/07}

\usepackage{lipsum}

\begin{document}

\title{A Length Adaptive Algorithm-Hardware Co-design of \\ Transformer on FPGA Through Sparse Attention and Dynamic Pipelining}



\author{Hongwu Peng$^{1,+}$, Shaoyi Huang$^{1,+}$, Shiyang Chen$^2$, Bingbing Li$^1$, Tong Geng$^3$, Ang Li$^3$, Weiwen Jiang$^4$, Wujie Wen$^5$, Jinbo Bi$^1$, Hang Liu$^2$ and Caiwen Ding$^1$}
\affiliation{$^+$These authors contributed equally. \country{}}
\affiliation{\fontsize{11pt}{11pt}\selectfont \institution{$^1$University of Connecticut \country{USA}. 
 $^2$Stevens Institute of Technology \country{USA}.  $^3$Pacific Northwest National Laboratory \country{USA}.  $^4$George Mason University \country{USA}.  $^5$Lehigh University \country{USA}. }
 \country{}}
\affiliation{\fontsize{9pt}{9pt}\selectfont
$^{1}$\{hongwu.peng, shaoyi.huang, bingbing.li, jinbo.bi, caiwen.ding\}@uconn.edu, 
$^{2}$\{schen94, hliu77\}@stevens.edu, \\
$^{3}$\{tong.geng, ang.li\}@pnnl.gov, $^{4}$wjiang8@gmu.edu, $^{5}$wuw219@lehigh.edu
\country{}
}

\renewcommand{\shortauthors}{Peng and Huang, et al.}

\begin{abstract}


Transformers are considered one of the most important deep learning models since 2018, in part because it establishes state-of-the-art (SOTA) records and could potentially replace existing Deep Neural Networks (DNNs). Despite the remarkable triumphs, the prolonged turnaround time of Transformer models is a widely recognized roadblock. The variety of sequence lengths imposes additional computing overhead where inputs need to be zero-padded to the maximum sentence length in the batch to accommodate the parallel computing platforms. This paper targets the field-programmable gate array (FPGA) and proposes a coherent sequence length adaptive algorithm–hardware co-design for Transformer acceleration. Particularly, we develop a hardware-friendly sparse attention operator and a length-aware hardware resource scheduling algorithm. 
The proposed sparse attention operator brings the complexity of attention-based models down to linear complexity and alleviates the off-chip memory traffic. The proposed length-aware resource hardware scheduling algorithm dynamically allocates the hardware resources to fill up the pipeline slots and eliminates bubbles for NLP tasks. Experiments show that our design has very small accuracy loss and has 80.2 $\times$ and 2.6 $\times$ speedup compared to CPU and GPU implementation, and 4 $\times$ higher energy efficiency than state-of-the-art GPU accelerator optimized via CUBLAS GEMM. 

\vspace{-2.5mm}

\end{abstract}





\keywords{
Transformer, Attention, BERT, Length adaptive, FPGA
\vspace{-1mm}
}


\maketitle


\vspace{-2mm}
\section{Introduction}

Transformers are considered as one of the most important deep learning models since 2018~\cite{multihead}, in part because it could potentially replace existing Deep Neural Networks (DNNs), such as Convolutional Neural Networks (CNN) and Recurrent Neural Networks (RNN)~\cite{cordonnier2019relationship}. By leveraging self-attention~\cite{vaswani2017attention}, Transformers have established state-of-the-art (SOTA) records (beyond human level) in various fields. Using computer vision as an example, CNNs were the first choice previously~\cite{he2016deep}; but nowadays, Transformer is gradually becoming the potential alternative, both from theoretical demonstration and empirical explorations, e.g., ~\cite{cordonnier2019relationship,dosovitskiy2020image,liu2021Swin,lin2021end-to-end}.

\begin{figure}[t]
    \includegraphics[width =0.98 \linewidth]{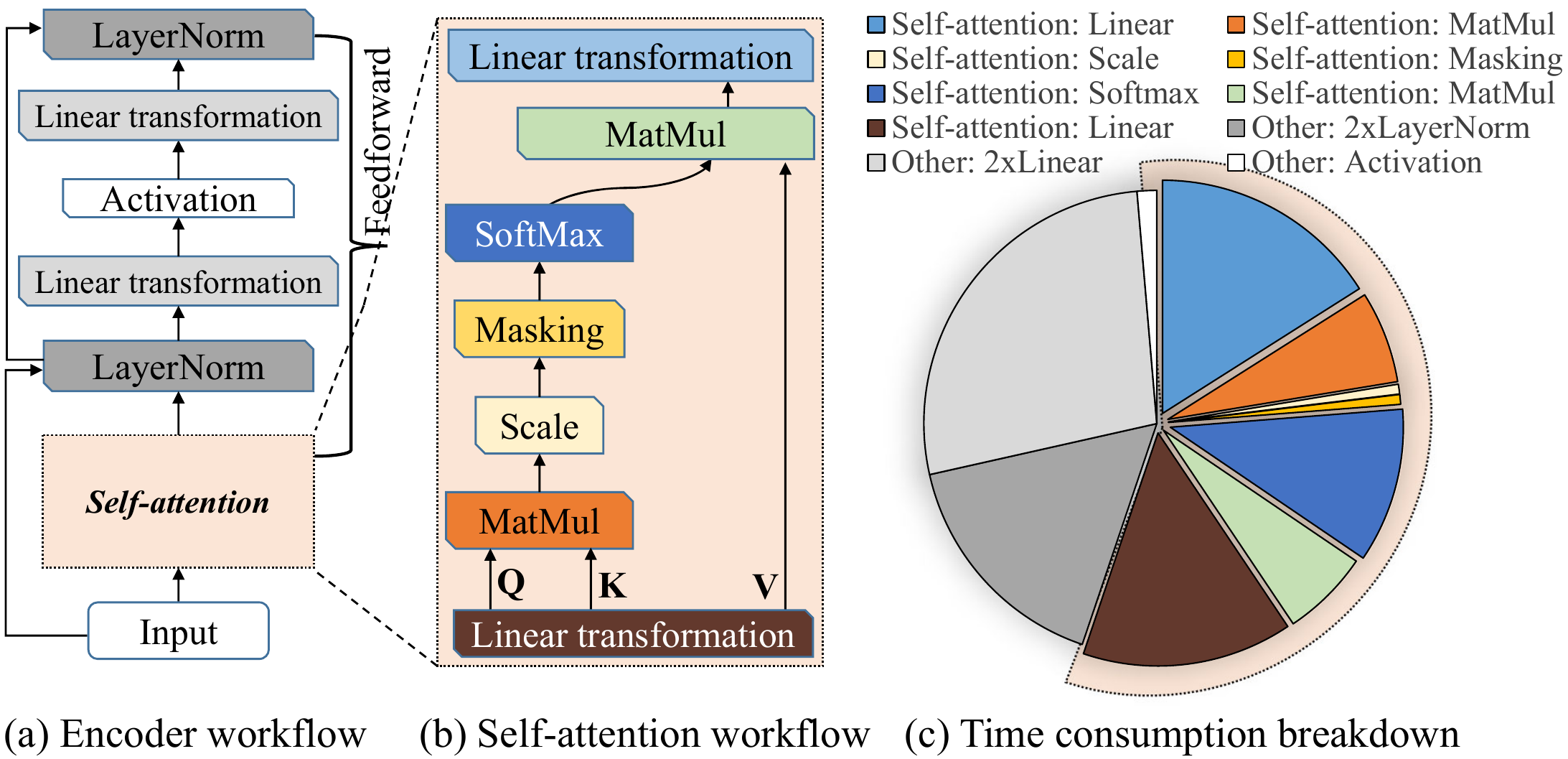} 
    \vspace{-3.5mm}
     \caption{The architecture of a four-head encoder. The time consumption is measured on for TensorRT~\cite{TensorRT} on WikiText-2 dataset~\cite{merity2016pointer}, where the input sequence has 128 tokens. 
     }
    \label{fig:encoder} 
    \vspace{-6mm}
\end{figure}

Despite the remarkable triumphs,
the prolonged turnaround time of Transformer models is a widely recognized roadblock that concerns real-world applications. Fig.s~\ref{fig:encoder}(a) and (b) depict the architecture of one encoder a.k.a. the building block for Transformers. Briefly, one encoder takes the word embeddings of a sequence as input. These embeddings pass through the self-attention mechanism to produce an attention matrix. This matrix is fed through layer normalization, linear transformation, and activation operations to derive the output. Various Transformer model variants often stack different numbers of encoders and decoders together~\cite{vaswani2017attention}. The bad news is that the time consumption of a single encoder in Fig.~\ref{fig:encoder} could easily reach 100s of $\mu$s, which is $\sim$10$\times$ slower than a typical CNN model. Of such a long latency, around 60\% of the time is spent in the self-attention workflow. According to our preliminary study, the time consumption ratio of self-attention is projected to climb if the number of tokens in the input sequence increases, especially in the NLP field~\cite {rajpurkar2018know} further.  
There exist several initial attention-aware optimization attempts, such as {sparse attention}~\cite{zaheer2020big} and {attention approximation}~\cite{ye2019bp,kitaev2019reformer,ham20203,wang2021spatten,ham2021elsa}, to leverage runtime approximations or domain knowledge, i.e., tokens only attend their nearby tokens \& a few sampled tokens as the summary of the sentence instead of attending all tokens. However, these explorations, unfortunately, fall short by either lacking generality or high computation overheads.

To make it worse, Transformer brings the challenge of a wide variety of input lengths, where inputs need to be zero-padded to the maximum sentence length in the batch to accommodate the parallel computing platforms such as GPU and Field-Programmable Gate Array (FPGA) \cite{wolf2020transformers}. By nature, RNN-based models, e.g., GRU and LSTM, process the inputs sequentially. Thus, the inputs could be divided into unified fixed-length sub-inputs and processed independently. Transformers leverages parallel processing and therefore cannot enjoy the benefit of fixed-length sub-inputs. Existing works on sequence length standardization fall into two categories. The first one is padding or truncation, which forces the sequence length to be the same. It leads to enormous computation overhead due to the unnecessary computation of the padding part. The second category divides a sequence batch into micro-batches (padding within the micro-batch) to mitigate the computation overhead. However, the various and irregular sequence length undermines overall performance and throughput at the inter micro-batch level.
Together with the prolonged turnaround time, achieving fast and efficient Transformer models becomes a grand challenge.

Across all the popular hardware, e.g., CPUs, GPUs, FPGAs, and Application-Specific Integrated Circuits (ASICs), FPGAs strike an effective balance among massive parallelism, high energy efficiency and short development cycle, hence lend themselves as the top choice to expedite the Transformer architecture. 
In this paper, we believe that the ideal Transformers acceleration should have a coherent algorithm--hardware co-design. We believe that \textbf{(i) Transformers should have their dedicated efficient algorithm designs}. Since self-attention cares more about the value relativity of all the attention scores than the absolute value of any specific attention score, we propose an efficient scheme to exploit two different self-attention approximations adaptively. Note, our approximation mechanisms are quantization-based designs that are not only computation-efficient but also hardware-friendly.
We also think that \textbf{(ii) Transformers should efficiently support various sequence length inputs}. For instance, {SQuAD v2.0} \cite{rajpurkar2018know} has an average and maximum sequence length of {171} and {975}, respectively. When padding the sequence with 975, it causes {5.7}$\times$ computational and memory bandwidth overhead on average. The inputs are sorted and processed according to the order of length. 
Compared to existing works, we achieve 4 $\times$ higher energy efficiency than GPU accelerator optimized  through CUBLAS GEMM routine \cite{chen2021re, huang2021hmc} with small accuracy loss, and comparable energy efficiency compared to ASIC accelerator designs \cite{ham20203, wang2021spatten}.
Our contributions are:
\begin{itemize}
\leftmargini=1mm
\item We propose sparse attention which is computation-efficient and hardware-friendly to reduce the need for computational resources and memory bandwidth. 
\item We propose a sequence length adaptive design to allocate coarse pipeline stages dynamically to eliminate pipeline bubbles and achieve the highest possible throughput under different sequence length inputs. 
\item 
Transformer exhibits a highly skewed distribution of computation complexity among the operators. We further develop a loop fusion to orchestrate the multiple attention operators and re-arrange various Transformer computations to enhance temporal locality and efficient hardware design with finer granularity.  
\end{itemize}

\vspace{-5mm}
\section{Related Work}

\noindent\textbf{Attention-aware Optimization.}
We also notice several recent attention-aware optimization attempts, such as \textit{sparse attention}~\cite{zaheer2020big} and \textit{attention approximation}~\cite{ye2019bp,kitaev2019reformer,ham20203,wang2021spatten,ham2021elsa}, which unfortunately fall short by either lack of generality or high computation overheads. Particularly, 
\textit{sparse attention mechanisms}~\cite{zaheer2020big} leverage domain knowledge, i.e., tokens only attend their nearby tokens and a few sampled tokens as the summary of the sentence instead of attending all tokens, to reduce computation and memory consumption during self-attention computation. Such design requires a pre-determined attention mask that lacks generality. 

\textit{Approximation-based attention} leverages runtime approximations to derive sparse attention which faces non-trivial overheads. BP-Transformer~\cite{ye2019bp} and Reformer~\cite{kitaev2019reformer} convert the self-attention computation into a nearest neighbor search problem and use either 
tree-based search, 
Locality Sensitive Hashing (LSH), or low-rank approximation
to find similar tokens for attention. 
A$^3$~\cite{ham20203} embraces architecture innovation to estimate the closeness between tokens. However, the estimation process still requires full access to original matrices and does not alleviate memory bottleneck for attention computation according to SpAtten~\cite{wang2021spatten}.
ELSA~\cite{ham2021elsa} uses LSH distance for attention rank approximation, but it again suffers significant overheads for LSH.

\noindent\textbf{Sequence length standardization.}
TensorRT~\cite{TensorRT} utilized the padding and truncation \cite{wolf2020transformers} to standardize the sequence length for parallel computing. It makes the hardware design regularized but leads to enormous computation overhead due to the unnecessary computation of the zero-padding part. TurboTransformer \cite{fang2021turbotransformers} divided a batch into micro-batches with similar lengths; however, within the micro-batch, it still required maximum sequence length padding. To make things worse, when we implement this method on FPGA, it introduces significant pipeline bubbles.


\vspace{-3.5mm}
\section{Sparse Attention Algorithm}
\vspace{-1mm}
\subsection{Overview}
\vspace{-1mm}

\textit{Approximation-based attention} leverages run-time approximations to derive sparse attention which faces non-trivial overheads. BP-Transformer~\cite{ye2019bp} and Reformer~\cite{kitaev2019reformer} convert the self-attention computation into a nearest neighbor search problem and use either tree-based search, Locality Sensitive Hashing (LSH), or low-rank approximation to find similar tokens for attention. A$^3$~\cite{ham20203} embraces architecture innovation to estimate the closeness between tokens. However, the estimation process still requires full access to original matrices and does not alleviate memory bottleneck for attention computation according to SpAtten~\cite{wang2021spatten}.


We propose to compute sparse self-attention via rapid attention rank approximation and sparse attention computation. We firstly quantize the full precision \textbf{Q} and \textbf{K} into low bits representation, i.e., 1 bit or 4 bits. Then we conduct matrix multiplication on quantized value and get the Top-$k$ candidates index. At last, we conduct full precision sparse attention computation based on Top-$k$ candidates. The algorithm \textbf{reduces the attention complexity from $O(n\textsuperscript{2})$ to $O(n)$}, where $n$ is sequence length.

\begin{figure*}[t]\vspace{-2mm}
\centering
\includegraphics[width=.95\textwidth]{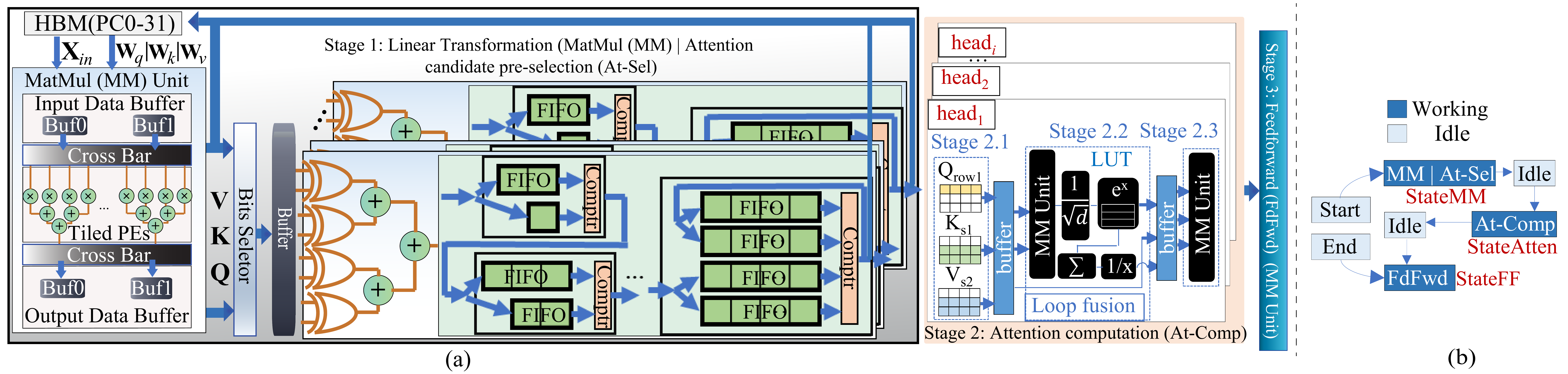}
\vspace{-3.5mm}
\caption{(a)
Sparse attention on FPGA; (b) State machine.}
\label{fig:hw_design}
\vspace{-4.5mm}
\end{figure*}

\vspace{-3.5mm}
\subsection{Sparse Attention Via \textbf{Q} \& \textbf{K} Quantization}
\label{subsec:graph}

\begin{figure}[t]
  \centering
  \includegraphics[width=0.9\linewidth]{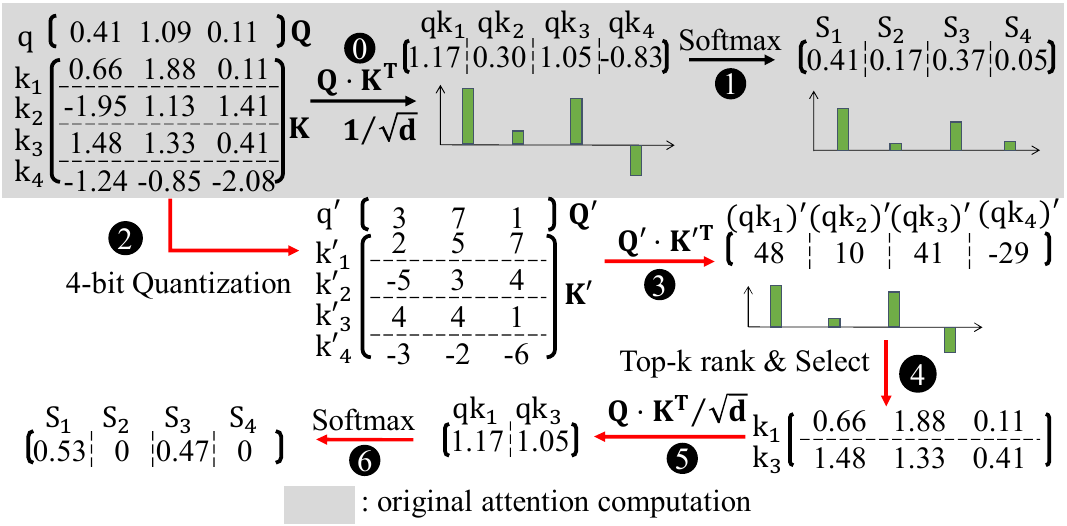}
  \vspace{-3mm}
  \caption{Candidate selection from quantized result.}
  \label{fig:softmax}
  \vspace{-6mm}
\end{figure}

In self-attention, the input is transformed into three matrices \textbf{Q}, \textbf{K} and \textbf{V}. Then \textbf{Q} and \textbf{K} are multiplied to arrive at $\mathbf{S} = \mathbf{Q} \cdot \mathbf{K}^T$, where each element in the resultant matrix is an attention score. A row of attention scores in \textbf{S} represent the dot-product between a row vector in \textbf{Q} and all row vectors in \textbf{K} respectively. 
Step \circled{0} in Fig.~\ref{fig:softmax} illustrates one row of \textbf{Q} multiplying with \textbf{K}, where $\mathbf{q}$ is one row in matrix \textbf{Q}. Subsequently, we perform a softmax operation on \textbf{S}, which is $\textbf{S}_{i} = \frac{\exp({\textrm{qk}_{i}})}{\sum^4_{j=1}\exp({\textrm{qk}_{j}})}$, i.e., step \circled{1} in Fig.~\ref{fig:softmax}. The key observation is: since softmax is a normalization method, \textit{it is the value relativity of all the attention scores, as opposed to the absolute value of any specific attention score, that matters.}

We propose to quantize \textbf{Q} and \textbf{K} from the full-precision representation (usually 32-bit floating-point) into a low-precision integer representation. 
Because both quantization and exponential operations used in softmax are monotonically increasing operators, the quantized results maintain the order of attention scores. We use our fast quantized matrix multiplication to extract dominant attention values. Afterward, we perform accurate attention computation only for dominant attention scores. The design is depicted in Fig.~\ref{fig:softmax}.

Particularly, we first find out the suitable scaling factor $M$ for the given tensor to quantization, then perform $x' = {round}(\frac{2^3-1}{|M|}x)$, which casts all the floating point values into a desired integer. For example, the scaling factor $M$ of \textbf{K} in Figure~\ref{fig:softmax} is $0.77$, so each element is be multiplied with $\frac{2^3-1}{0.77}$ and rounded to the nearest integer.
We follow a similar procedure to quantize q into q'. Subsequently, we again use a look-up table to perform the multiplication. For instance, if we multiply two 4-bit integers, the look-up table only needs 256 entries. We can easily estimate the multiplied value. At the end of step \circled{2}, we derive the $\mathbf{Q'} \cdot \mathbf{K'}^T$. As the examples indicate, the quantized results keep the same rank and distribution compared with their full-precision counterpart. 

We conduct Top-$k$ sort and select the Top-$k$ ranked attention scores for exact matrix multiplication, which derives more accurate softmax values. This is faster than the original design because we only need to compute Top-$k$ attention scores. In step \circled{4} in Fig.~\ref{fig:softmax}, we select Top-2 element $k_1$ and $k_3$ to perform matrix multiplication and softmax, which is used as an approximation of the result of self-attention. Subsequently, we will perform full-precision  $\mathrm{\bf Q} \cdot \mathrm{\bf K}^T$ for the selected attention scores at \circled{5} and final softmax at step \circled{6}.

\vspace{-2.5mm}
\section{Hardware Accelerator Design and Scheduling Algorithm}

The proposed quantization-based sparse attention system design is more fit on FPGAs than general-purpose processors because the latter are instruction-driven architecture. At the same time, FPGAs are data-driven architecture that avoids instruction fetch and related memory access. When compared to the popular GPU accelerators, FPGAs excel for the following reasons:
(1) The on-chip memory capacity of FPGAs is much higher (360$\times$) than that of GPUs (i.e., 35 MB in Xilinx Alveo U200 vs. 96 KB in V100). 
The FPGA on-chip memory features its high memory bandwidth (31 TB/s) and low access latency (single clock cycle), enabling higher throughput and lower latency design \cite{peng2020selective, wang2020benchmarking, qi2021accommodating, peng2021binary, peng2021accelerating, yuan2021improving, huang2022automatic, peng2020design}.
With more on-chip memory size, we can achieve a better computation to communication (CTC) ratio for the same operations, i.e., matrix multiply and matrix add. 
(2) FPGA provides more design opportunities on fine-grained and coarse-grained pipelining and loop fusion techniques. We can have better data locality optimization, and design space freedom on FPGA through polyhedral analysis and proper loop reschedule. 
(3)  FPGAs are more power-efficient, and therefore more suitable for resource-constrained scenarios. 

The FPGA platform enables better intra-attention coarse grain pipelining design and leaves more freedom on FPGA resource allocation. The commonly used way for NLP tasks with variable length inputs is to unify input sequence to a fixed length through padding and cutting. However, length padding introduces unnecessary overhead, and length cutting leads to information loss.
To accelerate the NLP tasks with variable sequence length, we propose sequence length adaptive Transformer hardware design on the FPGA platform and the corresponding hardware scheduling algorithms to optimize the coarse-grained stage throughput. Compared to commonly used padding and cutting methods, our proposed method has higher hardware throughput and less information loss. 

\vspace{-3mm}
\subsection{Sparse Attention Accelerator Design}

We break down the original single Transformer Encoder stage pipeline into three coarse-grained pipelines and overlap their execution time by inserting buffers for each concatenated pipeline pair, as shown in Fig.~\ref{fig:hw_design}(a).
Stage 1 contains the linear transformation (using MatMul (MM))
and quick attention approximations (a.k.a, pre-selection) (At-Sel) 
hardware. 
The MM result is directly fed into the bits selector hardware for ultra-low bit quantization. The result is stored in the on-chip buffer for candidate pre-selection computation (utilizing LUT hardware for approximate distance calculation).
The approximate distance output and address are then fed to the merge sort hardware for high throughput (II=1) scalable Top-$k$ sort \cite{peng2021optimizing}. The Top-$k$ results (e.g., index and value pairs) are stored back to HBM for inter-stage buffering. 
Stage 2 is attention computation (At-Comp) and Stage 3 is feedforward (FdFwd).
Stage 2 is divided into three sub-stages and implemented with the intra-layer coarse-grained pipeline to enhance hardware utilization. 
Stage 2.1, the data loading stage, utilizes the Top-$k$ results from stage 1 to choose the attention candidates.
As discussed in Section.~\ref{subsec:graph}, each $Q_{rowi}$ have selected candidates $K_{si}$ and $V_{si}$ matrix for attention computation. 
Stage 2.2 is implemented with loop fusion and is composed of operators \circled{4} to \circled{6} in Fig.~\ref{fig:softmax}. Stage 2.3 is composed of $Z_i = S_i \cdot V/sum(S_i)$ operation for each row $i$. 
The double-buffers added between stages buffer the data produced/consumed by the previous/current stage for coarse-grained pipelining. Stage 2.3 conduct the final MM operation between $Z_i$ and $V$
Stage 3 (feedforward) is composed of MM, add, layer normalization, and GELU unit.

\begin{figure}[t]
  \centering
  \includegraphics[width=0.9\linewidth]{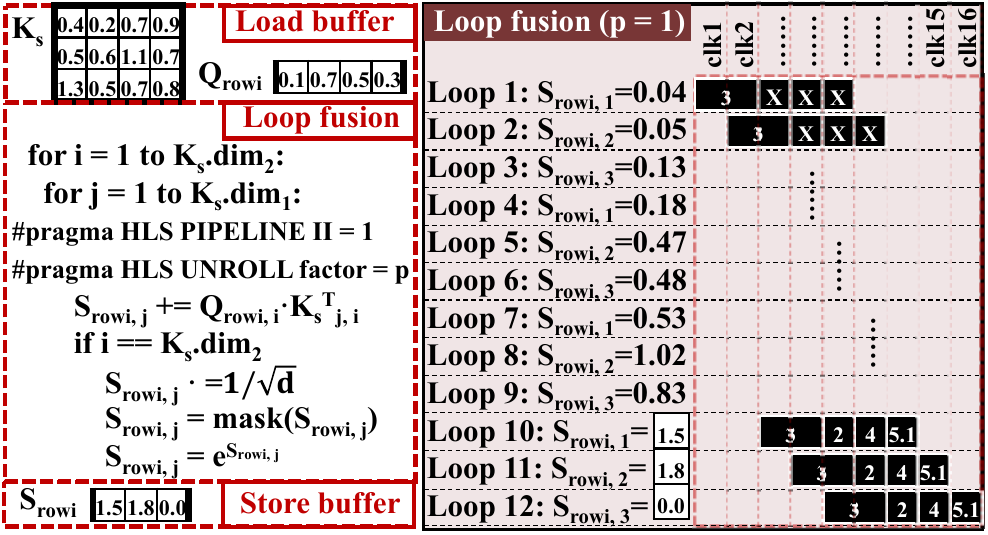}
  \vspace{-4mm}
  \caption{An attention kernel fusion example.}
  \label{fig:loopfusion}
  \vspace{-7.5mm}
\end{figure}

In stage 2, We divide the softmax \circled{6} in Fig.~\ref{fig:softmax} into two operations: exponent calculation and normalization. 
We leverage the FPGA's fine-grained pipelining characteristic to fuse multiple attention operators
(i.e., \circled{5} to \circled{6.1}) 
into a single loop.  An example is given in Fig.~\ref{fig:loopfusion}, the scaling, mask, and exponential operations are conducted at the last loop iterations. 
$Q_{rowi}$ is the processed row and $K_s$ is the corresponding selected candidate for the attention. $K_s.dim_1$ and $K_s.dim_2$ are the size of $1\textsuperscript{st}$ and $2\textsuperscript{st}$ dimension of $K_s$. $S_{rowi}$ is the processed attention score result. 
Thanks to the reconfigurable architecture of FPGAs, we enable fusing the loops with different iteration trip counts while GPU designs (e.g., TensorRT) only can fuse specific loops (e.g., scale + matrix multiply). Both Stage 1 and 3 leverage on-chip memory and proper loop allocation to mitigate communication bottlenecks between the on and off-chip memory. 

\vspace{-2mm}
\subsection{Length-aware Scheduling Algorithm}
\label{sec:length_aware}

\begin{figure*}[t]\vspace{-2mm}
\centering
\includegraphics[width=.99\textwidth]{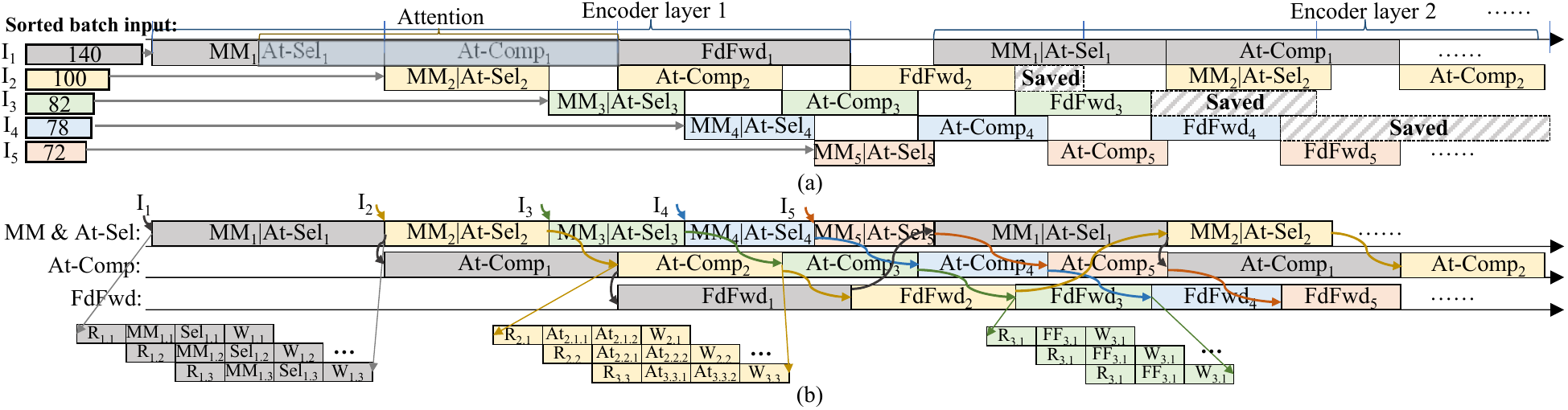}
\vspace{-3.5mm}
\caption{Length-aware coarse-grained dynamic pipeline algorithm example: (a) timing diagram; (b) hardware utilization.}
\label{fig:Timing_diagram}
\vspace{-4.5mm}
\end{figure*}

Different length sequences tasks consume different computational resources and have different latency, which introduces pipeline bubbles due to irregular and unpredictable dataflow. Because all operators have $O(n)$ complexity where n is sequence length, we proposed a novel length-aware coarse-grained pipeline algorithm to dynamically adjust the hardware resource allocation to illuminate the pipeline bubble. We leverage FPGA characteristics to adjust the resource allocation according to stages' computation complexity, eliminate redundant computation and achieve ultra-high throughput. 
We propose the techniques detailed below.

\textbf{Resource Scheduling Algorithm. } 
The slowest stage constrains the throughput of the coarse-grained pipeline. We first develop a Encoder coarse-grained stage allocation algorithm (Algorithm~\ref{algorithm:coarse_grain}) to schedule operators efficiently. For original operator graph $G=(V,E)$, each vertex $v_i\in V$ represents an operator and the edge $e_{ij}$ represents the data dependency between $v_i$ and $v_j$. Each vertex $v_i$ has a weight $W(w_i, s_{avg})$ which is the associated arithmetic computational complexity. 
It takes the $G$ and Encoder operator weight set $W(V, s_{avg})$ at average sequence length $s_{avg}$, and Encoder operator priority set at average sequence length $P(V, s_{avg})$ as input and outputs operator subgraph of each Encoder computation stage $G_k = (V_k, E_k)$ shown in Eq.~\ref{eq:priority}, where $W(v_i, s_{avg})$ and $P(v_i, s_{avg})$ denote their value at vertex ($v_i$, $s_{avg}$). To fully utilize the resources of a certain FPGA chip for sequence length adaptive design, we further adjust the operator parallelism $N(v_i, s_i)$ for intra coarse-grained pipeline stages and enumerate pipeline replication factor $R(G_k, s_i)$  to obtain the optimal setting with the help of analytical performance and resource models. 
\vspace{-.06in}
 \begin{equation} \label{eq:priority} \small
  P(v_i, s_{avg}) = \begin{cases} W(v_i, s_{avg}) + \max\limits_{v_j\in Succ(v_i)}{P(v_j, , s_{avg})},\ v_i \neq v_{sink} \\
 W(v_{sink}, s), \ \ \ \ \ \ \ \ \ \ \ \ \ \ \ \ \ \ \ \ \ \ \ \text{otherwise}
  \vspace{0em}
\end{cases}
\end{equation} 


 
  \begin{algorithm}[t]
\scriptsize
 \KwIn{Encoder operator graph $G = (V, E)$, Encoder operator weight set $W(V, s_{avg})$, and Encoder priority set $P(V, s_{avg})$;}
 \KwOut{operator subgraph of each Encoder computation stage $G_k = (V_k, E_k)$;}
 Traverse $G=(V,E)$ and compute priority set $P(V, s_{avg})$ for the Encoder;\\
 $k \leftarrow 1, N(V) \leftarrow \{\textbf{1}\}$, $G_1 \leftarrow v_1$; \tcp{add a operator to a new stage} 
 \ForEach{$v_i \in$ V in decreasing order of $P(v, s_{avg})$}{
        \ForEach{$N'(v_j) \in G_k$}{
            $N'(v_j) \leftarrow N(v_j)\cdot \ceil{\frac{W(v_j, s_{avg})}{W(v_i, s_{avg})}}$;
        }
    \uIf{resource constraints are satisfied} {
    $G_j \leftarrow v$; \tcp{add a operator to current stage}
    $N(V) \leftarrow N'(V)$; \tcp{update operator parallelisms}
    } 
    \Else{$k, K \leftarrow k+1$;  \\
    $G_k \leftarrow v_i$; \tcp{add a operator to a new stage}}
     }
 \Return{\{$G_1, G_2, ..., G_K$\}};\\ 
 \caption{Encoder coarse-grained Stage Allocation.}\label{algorithm:coarse_grain}
\end{algorithm}
 \vspace{0mm}


\textbf{Length-aware Coarse-grained Pipeline Algorithm.} 
We then develop a length adaptive resource scheduling method to dynamically patch the pipeline bubbles for a batch of tasks with different sequence lengths. The effectiveness of the proposed scheduling method relies on the fact that \textbf{all operators have O(n) complexity}. 
The batch inputs are sorted and processed according to the decreasing order of length, under the control of a dedicated state machine (three states shown in Fig.~\ref{fig:hw_design}(b): $StateMM$, $StateAtten$ and $StateFF$) during its Encoder activation period. 
The state machines dynamically allocate hardware resources (stages and buffers) to eliminate pipeline bubbles and ensure a high hardware utilization for batch sentences with varying lengths. 
Each stage has almost 100\% utilization, and there is no pipeline bubble. We could significantly reduce the latency (denoted as "saved"). 

A length-aware scheduling timing diagram example is given in Fig.~\ref{fig:Timing_diagram}, where the batch size is 5, and the input sequence length varies from 72 to 140. The batch inputs are sorted and fed into the Encoder coarse-grained stages in decreasing order of sequence length. Fig.~\ref{fig:Timing_diagram}(a) shows how each of the Encoder coarse-grained stages processes each sequence input. Most attention-based models have multiple Encoder layers, so the batch input is processed by the layer order. Fig.~\ref{fig:Timing_diagram}(b) shows the hardware resource occupation of Encoder coarse-grained stages. With the state machine-based scheduling algorithm implemented, the pipeline stages of different sequence length inputs and different Encoder layers are patched together without pipeline bubble, so both stages have almost 100\% hardware utilization. The intra-layer coarse-grained pipeline is implemented in each stage to exploit the trade-off between spatial \& temporal data locality and hardware resource occupation. The communication and computation are overlapped with each other through coarse-grained pipeline and data prefetching. 

\vspace{-2mm}
\section{Experiment}

We evaluate several well-known self-attention centric models to demonstrate the algorithm \& hardware design performance. For NLP models, we choose four of the most popular ones: BERT-base \cite{devlin2018bert}, BERT-large, and DistilBERT \cite{sanh2019distilbert}, and RoBERTa \cite{liu2019roberta}. 
Model configurations are given in Table~\ref{table:Model}. These models are self-attention-centric and have a similar structure. For BERT-base, DistilBERT and RoBERTa, we run 3 representative datasets to evaluate the performance: SQuAD v1.1 \cite{rajpurkar2016squad}, RTE \cite{dagan2010recognizing}, and MRPC \cite{dolan2005automatically}. For BERT-large, we run SQuAD v1.1 for evaluation. 
The minimum sequence length, average sequence, and maximum sequence of those datasets are also given. Max/Avg ratio also corresponds to the computational overhead introduced through padding.





\begin{table}[htbp]   \small
\vspace{-4mm}
    \centering
    \caption{Model \& evaluation dataset. }
     \label{table:Model}
    \vspace{-3.5mm}
    \begin{tabular}{rrrrr}
    \hline
    \multicolumn{1}{|c|}{Model} & \multicolumn{1}{c|}{Layers} & \multicolumn{1}{c|}{Hidden dim} & \multicolumn{1}{c|}{Num. of Heads}\\
    \hline 
    \multicolumn{1}{|c|}{DistilBERT} & \multicolumn{1}{c|}{6} & \multicolumn{1}{c|}{768} & \multicolumn{1}{c|}{12}\\
    \hline 
    \multicolumn{1}{|c|}{BERT-base, RoBERTa} & \multicolumn{1}{c|}{12} & \multicolumn{1}{c|}{768} & \multicolumn{1}{c|}{12}\\
    \hline 
    \multicolumn{1}{|c|}{BERT-large} & \multicolumn{1}{c|}{24} & \multicolumn{1}{c|}{1024} & \multicolumn{1}{c|}{16}\\
    
    \hline 
    \hline 
    \multicolumn{1}{|c|}{Evaluation dataset} & \multicolumn{1}{c|}{Avg} & \multicolumn{1}{c|}{Max} & \multicolumn{1}{c|}{Max/Avg}\\
    \hline 
    \multicolumn{1}{|c|}{SQuAD v1.1} & \multicolumn{1}{c|}{177} & \multicolumn{1}{c|}{821}& \multicolumn{1}{c|}{4.6}\\
    \hline 
    \multicolumn{1}{|c|}{RTE} &  \multicolumn{1}{c|}{68} & \multicolumn{1}{c|}{253}& \multicolumn{1}{c|}{3.7}\\
    \hline 
    \multicolumn{1}{|c|}{MRPC} &  \multicolumn{1}{c|}{53} & \multicolumn{1}{c|}{86}& \multicolumn{1}{c|}{1.6}\\
    \hline 
    \vspace{-7mm}
    \end{tabular}
\end{table}

The FPGA hardware design and evaluation are conducted on the Alveo U280 platform. We also evaluate the hardware performance on CPU, edge GPU, and GPU server platforms for cross-platform comparison: Intel(R) Xeon(R) Gold 5218 CPU, Jetson TX2, and Quadro RTX 6000. The FPGA design is conducted on software version Vivado 2020.1, whereas GPU and CPU design is completed on Pytorch 1.10.0 and Transformers 4.13.0.dev0.


\vspace{-4mm}
\subsection{Sparse Attention Accuracy Evaluation}

We evaluate the models mentioned above and datasets on the Top-$k$ sparse attention algorithm. The state-of-the-art models are quantized into 8 bits fixed-point representation without accuracy drop \cite{zhang2020ternarybert}. 
The $Q$ \& $K$ quantization is conducted based on 1-bit quantization, which is a sign function. 
For the sparse attention algorithm, the quantized models are directly used without model fine-tuning. 
For SQuAD v1.1 and MRPC datasets, we use the F1 score as our accuracy measure. For the RTE dataset, the raw accuracy is reported. Experimented are conducted on the corresponding validation dataset. 

\begin{figure*}[t]
\centering
\includegraphics[width=.99\textwidth]{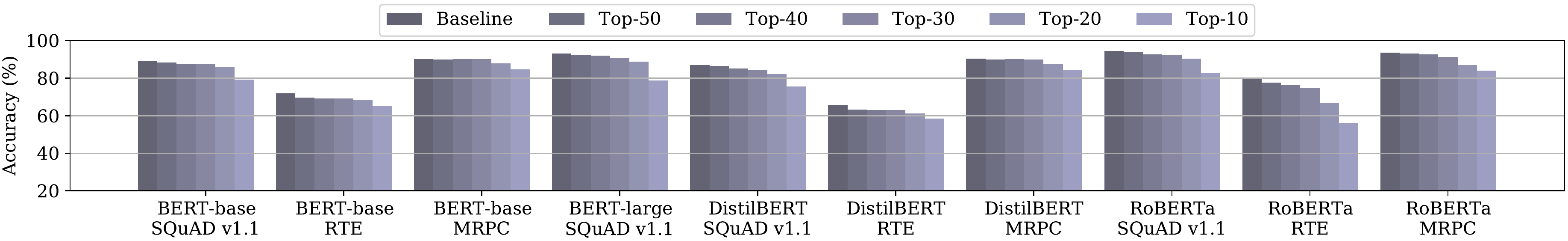}
\vspace{-4mm}
\caption{Accuracy evaluation of Top-$k$ sparse attention.}
\label{fig:sparse_aten}
\end{figure*}

Fig.~\ref{fig:sparse_aten} shows the accuracy test of evaluated models and datasets. We choose $k$ value from 10 to 50 to assess the effectiveness of sparse attention, where the $k$ value determines the degree of approximation for sparse attention computation. Generally, smaller $k$ indicates aggressive approximation and leads to a higher accuracy drop. For most of the evaluation, Top-10  sparse attention lead to non-negligible performance degradation. Top-30 provides a good trade-off between the accuracy drop and sparsity ratio, whereas all evaluations have less than 2\% accuracy drop. With a Top-30 sparse attention, the attention computation complexity can be reduced by more than 80\% in average. 


\begin{figure*}[h!]\vspace{-10mm}
    \centering
    \centering
\begin{multicols}{2}
\subfloat [\label{fig:hardware_throughput}End to end cross platform throughput comparison]   {\includegraphics[width=1.\columnwidth]{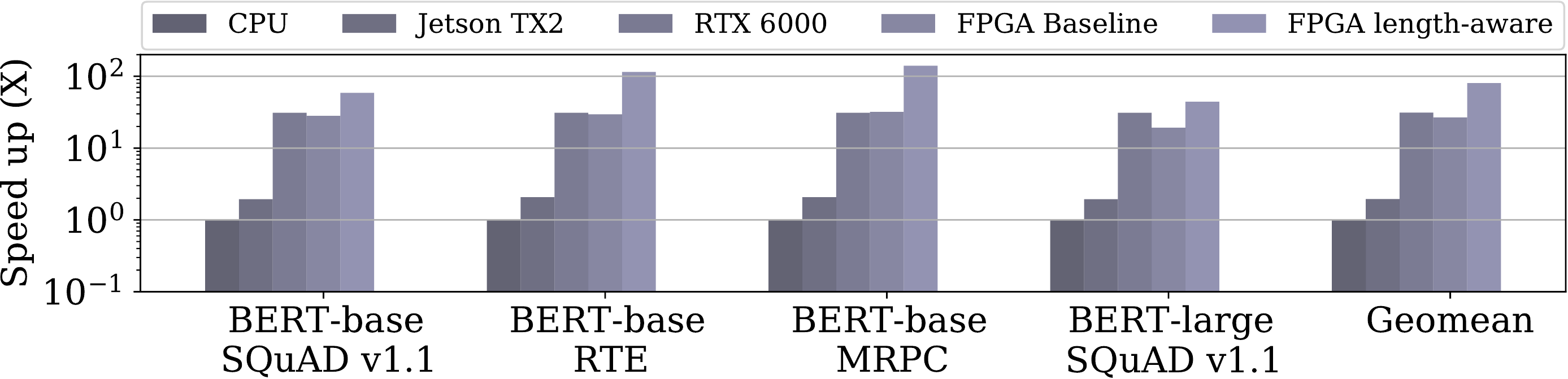}\par }
\subfloat  [\label{fig:attention_throughput}Cross platform attention throughput comparison]  {\hspace{.3in}\includegraphics[width=1.\columnwidth]{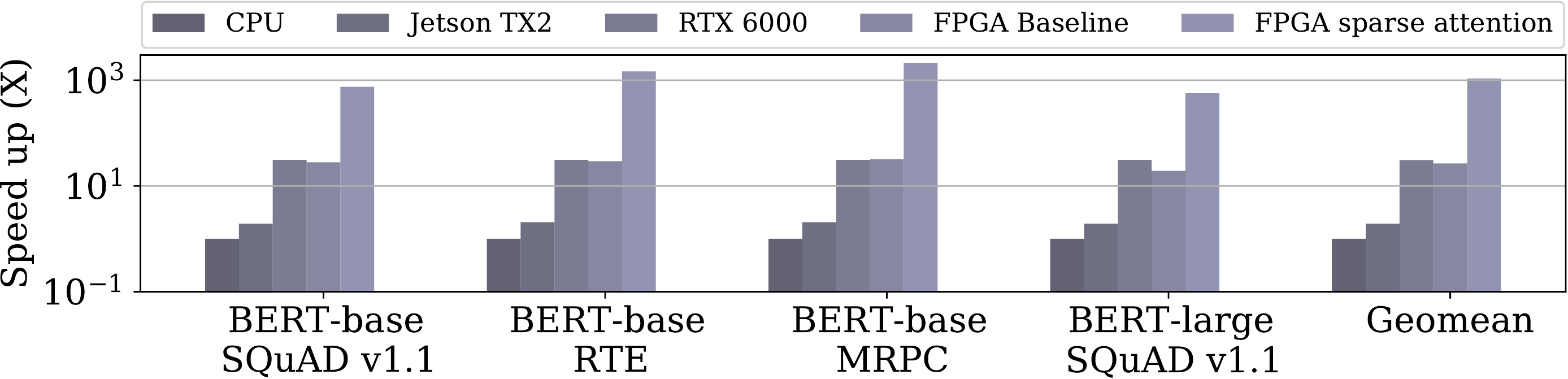}\par }
\end{multicols}
 \vspace{-6.5mm}
    \caption{Cross platform hardware evaluation.}
    \label{fig:hardware}
     \vspace{-3mm}
\end{figure*}






\vspace{-4mm}
\subsection{Cross-platform Throughput Evaluation}

Based on the model accuracy evaluation, we choose a sweet point $k = 30$ for Top-$k$ pre-selection, then we mapped models into FPGA hardware coarse-grained stages through Algorithm~\ref{algorithm:coarse_grain}. We exploit the design space to maximize the hardware throughput and CTC ratio for the hardware design. The attainable FPGA design frequency is 200 MHz, and most of the hardware resources (BRAM, FF, LUT) consumption are congested inside the SLR0 of the Alveo U280 board since the only SLR0 is connected to HBM channels. HBM channels provide a maximum of 460 GB/s bandwidth. 
The batch size is set as 16 to maximize the hardware utilization. 

For the FPGA platform, 8 bits fixed-point number multiply \& accumulate consumes 1 DSP unit. And there are
3000 DSP units within the SLR0 in total. 
So the maximum attainable computation throughput of the FPGA platform is 1.2 TFLOPS. Thanks to the reconfiguration structure, the FPGA design can efficiently map variable sequence length computation into hardware through the scheduling algorithm discussed in Sec.~\ref{sec:length_aware}. The FPGA design can surpass GPU server performance in all the models \& tasks evaluation through efficient scheduling. 
BERT-base on SQuAD v1.1, RTE, and MRPC, and BERT-large model on SQuAD v1.1 are used for hardware design demonstration. DistilBERT and RoBERTa has similar structure and thus similar hardware performance to BERT-base. 

The end-to-end hardware throughput comparison is given in Fig.~\ref{fig:hardware_throughput}. FPGA baseline indicates the FPGA design without length-ware scheduling and sparse attention algorithm implemented. The sequence length is padded to the maximum sequence length for the CPU and GPU design to evaluate the tasks. The geomean speedup of the FPGA length-aware sparse attention design is 80.2 $\times$, 41.3 $\times$, 2.6 $\times$, 3.1 $\times$ than CPU, edge GPU, GPU server, and FPGA baseline design. As for the self-attention computation, the hardware throughput is also recorded during the evaluation, and the corresponding speedup is given in  Fig.~\ref{fig:attention_throughput}. The FPGA sparse attention hardware achieves a geomean speedup of 1073 $\times$, 550 $\times$, 35 $\times$, 41 $\times$ than CPU, edge GPU, GPU server, and FPGA baseline design. 
Our FPGA design achieves an equivalent hardware throughput of 3.6 TFLOPS on 8 bits fixed-point operations with length-aware scheduling and sparse attention algorithm implemented.

\vspace{-2mm}
\subsection{Cross-work Energy Efficiency Comparison}


We further conduct cross-platform and cross-work energy efficiency comparison between GPU baseline, GPU design optimized through CUBLAS GEMM routine \cite{chen2021re}, FPGA \cite{qi2021accelerating}, and ASIC implementations \cite{ham20203, wang2021spatten} of Transformer accelerator. The result is shown in Table~\ref{table:energy}. Our FPGA surpasses GPU baseline implementation and existing state-of-the-art FPGA design in terms of both throughput and energy efficiency with an acceptable accuracy drop. With the length-aware scheduling and sparse attention algorithms implemented, our work has a comparable energy efficiency to existing ASIC designs dedicated to transformer acceleration. 
\begin{table}[htbp]   \small
    \centering
    \caption{Energy efficiency \& throughput comparison.}
     \label{table:energy}
    \vspace{-2mm}
    \begin{tabular}{rrrrr}
    \hline
    \multicolumn{1}{c|}{Work/platform} & \multicolumn{1}{c|}{Throughput} & \multicolumn{1}{c|}{Energy eff.} & \multicolumn{1}{c}{Accuracy drop}\\
    \multicolumn{1}{c|}{ } & \multicolumn{1}{c|}{(GOPS)} & \multicolumn{1}{c|}{(GOP/J)} & \multicolumn{1}{c}{(average)(\%)}\\
    \hline 
    \multicolumn{1}{c|}{GPU RTX 6000} & \multicolumn{1}{c|}{1380} & \multicolumn{1}{c|}{8} & \multicolumn{1}{c}{1.8}\\
    \hline 
    \multicolumn{1}{c|}{GPU V100: E.T. \cite{chen2021re}} & \multicolumn{1}{c|}{7550} & \multicolumn{1}{c|}{25} & \multicolumn{1}{c}{2.1}\\
    \hline 
    \multicolumn{1}{c|}{\textbf{Ours FPGA}} & \multicolumn{1}{c|}{3600} & \multicolumn{1}{c|}{102} & \multicolumn{1}{c}{1.8}\\
    \hline 
    
    \multicolumn{1}{c|}{FPGA design \cite{qi2021accelerating}} & \multicolumn{1}{c|}{76} & \multicolumn{1}{c|}{N/A} & \multicolumn{1}{c}{3.8}\\
    \hline 
    \multicolumn{1}{c|}{ASIC: A\textsuperscript{3} \cite{ham20203}} & \multicolumn{1}{c|}{221} & \multicolumn{1}{c|}{269} & \multicolumn{1}{c}{1.6}\\
    \hline 
    \multicolumn{1}{c|}{ASIC: SpAtten \cite{wang2021spatten}} & \multicolumn{1}{c|}{360} & \multicolumn{1}{c|}{382} & \multicolumn{1}{c}{1.1}\\
    \hline 

    \vspace{-2mm}
    \end{tabular}
\end{table}
    
\section{Conclution}
\vspace{-1mm}

In this paper, we propose a hardware-friendly sparse attention algorithm through query and key values quantization, where we bring down the complexity of self-attention from $O(n\textsuperscript{2})$ to $O(n)$.  We develop a length-aware hardware scheduling algorithm to accommodate variable sequence length computation into coarse-grained pipeline stages without pipeline bubbles. To alleviate the off-chip memory access, we further develop an attention kernel fusion to process the attention computation. 
We exploit temporal data locality through the on-chip buffer to enhance the CTC ratio and push the hardware design to the computation roof. Experimental results show that we achieve 80.2 $\times$  and 2.6 $\times$ speedup compared to CPU and GPU implementation with 1.8\% accuracy loss. Our FPGA design has more than 4 $\times$ higher energy efficiency than GPU accelerator optimized through CUBLAS GEMM routine and has comparable energy efficiency compared to state-of-the-art ASIC designs.

\section*{Acknowledgement}
This work  was  in  part  supported  by the NSF CRII Award No. 2000722, CAREER Award No. 2011236, No. 2006748, No. 2046102, and DOE Award No. 66150. Any opinions, findings and conclusions, or recommendations expressed in this material are those of the authors and do not necessarily reflect the views of the funding agencies.

\vspace{-1mm}

\bibliography{ref}

\begin{thebibliography}{10}

\bibitem{multihead}
Nikolas Adaloglou.
\newblock {Why Multi-head Self Attention Works: Math, Intuitions and 10+1
  Hidden Insights}.
\newblock \url{https://theaisummer.com/self-attention/}, 2021.
\newblock [Online; accessed August 30, 2021].

\bibitem{cordonnier2019relationship}
Jean-Baptiste Cordonnier, Andreas Loukas, and Martin Jaggi.
\newblock {On the Relationship Between Self-Attention and Convolutional
  Layers}.
\newblock In {\em ICLR}, 2019.

\bibitem{vaswani2017attention}
Ashish Vaswani et~al.
\newblock {Attention is All You Need}.
\newblock In {\em Advances in neural information processing systems}, pages
  5998--6008, 2017.

\bibitem{he2016deep}
Kaiming He, Xiangyu Zhang, Shaoqing Ren, and Jian Sun.
\newblock {Deep Residual Learning for Image Recognition}.
\newblock In {\em CVPR}, 2016.

\bibitem{dosovitskiy2020image}
Alexey Dosovitskiy, Lucas Beyer, Alexander Kolesnikov, Dirk Weissenborn,
  Xiaohua Zhai, Thomas Unterthiner, Mostafa Dehghani, Matthias Minderer, Georg
  Heigold, Sylvain Gelly, Jakob Uszkoreit, and Neil Houlsby.
\newblock {An Image is Worth 16$\times$ 16 Words: Transformers for Image
  Recognition at Scale}.
\newblock In {\em ICLR}, 2021.

\bibitem{liu2021Swin}
Ze~Liu, Yutong Lin, Yue Cao, Han Hu, Yixuan Wei, Zheng Zhang, Stephen Lin, and
  Baining Guo.
\newblock {Swin Transformer: Hierarchical Vision Transformer using Shifted
  Windows}.
\newblock In {\em CVPR}, 2021.

\bibitem{lin2021end-to-end}
Kevin Lin, Lijuan Wang, and Zicheng Liu.
\newblock {End-to-End Human Pose and Mesh Reconstruction with Transformers}.
\newblock In {\em CVPR}, 2021.

\bibitem{rajpurkar2018know}
Pranav Rajpurkar, Robin Jia, and Percy Liang.
\newblock {Know What You Don't Know: Unanswerable Questions for SQuAD}.
\newblock In {\em ACL}, pages 784--789, 2018.

\bibitem{zaheer2020big}
Manzil Zaheer, Guru Guruganesh, Kumar~Avinava Dubey, Joshua Ainslie, Chris
  Alberti, Santiago Ontanon, Philip Pham, Anirudh Ravula, Qifan Wang, Li~Yang,
  and Amr Ahmed.
\newblock {Big Bird: Transformers for Longer Sequences}.
\newblock In H.~Larochelle, M.~Ranzato, R.~Hadsell, M.~F. Balcan, and H.~Lin,
  editors, {\em NeurIPS}, volume~33, pages 17283--17297. Curran Associates,
  Inc., 2020.

\bibitem{ye2019bp}
Zihao Ye, Qipeng Guo, Quan Gan, Xipeng Qiu, and Zheng Zhang.
\newblock {BP-Transformer: Modelling Long-range Context via Binary
  Partitioning}.
\newblock {\em arXiv preprint arXiv:1911.04070}, 2019.

\bibitem{kitaev2019reformer}
Nikita Kitaev, Lukasz Kaiser, and Anselm Levskaya.
\newblock {Reformer: The Efficient Transformer}.
\newblock In {\em ICLR}, 2019.

\bibitem{ham20203}
Tae~Jun Ham, Sung~Jun Jung, Seonghak Kim, Young~H Oh, Yeonhong Park, Yoonho
  Song, Jung-Hun Park, Sanghee Lee, Kyoung Park, Jae~W Lee, and Deog-Kyoon
  Jeong.
\newblock {A\^{} 3: Accelerating Attention Mechanisms in Neural Networks with
  Approximation}.
\newblock In {\em 2020 HPCA}, pages 328--341. IEEE, 2020.

\bibitem{wang2021spatten}
Hanrui Wang, Zhekai Zhang, and Song Han.
\newblock {SpAtten: Efficient Sparse Attention Architecture with Cascade Token
  and Head Pruning}.
\newblock In {\em 2021 HPCA}, pages 97--110. IEEE, 2021.

\bibitem{ham2021elsa}
Tae~Jun Ham, Yejin Lee, Seong~Hoon Seo, Soosung Kim, Hyunji Choi, Sung~Jun
  Jung, and Jae~W Lee.
\newblock {ELSA: Hardware-Software Co-design for Efficient, Lightweight
  Self-Attention Mechanism in Neural Networks}.
\newblock In {\em ISCA}. IEEE, 2021.

\bibitem{TensorRT}
NVIDIA.
\newblock {TensorRT}.
\newblock Retrived from \url{https://developer.nvidia.com/tensorrt}.
\newblock Online; accessed: October 6, 2021.

\bibitem{merity2016pointer}
Merity Stephen, Xiong Caiming, Bradbury James, and Socher Richard.
\newblock {Pointer Sentinel Mixture Models}.
\newblock ICLR, 2016.

\bibitem{wolf2020transformers}
Thomas Wolf, Julien Chaumond, Lysandre Debut, Victor Sanh, Clement Delangue,
  Anthony Moi, Pierric Cistac, Morgan Funtowicz, Joe Davison, Sam Shleifer,
  et~al.
\newblock Transformers: State-of-the-art natural language processing.
\newblock In {\em EMNLP}, pages 38--45, 2020.

\bibitem{chen2021re}
Shiyang Chen, Shaoyi Huang, Santosh Pandey, Bingbing Li, Guang~R Gao, Long
  Zheng, Caiwen Ding, and Hang Liu.
\newblock Et: re-thinking self-attention for transformer models on gpus.
\newblock In {\em Proceedings of the International Conference for High
  Performance Computing, Networking, Storage and Analysis}, pages 1--18, 2021.

\bibitem{huang2021hmc}
Shaoyi Huang, Shiyang Chen, Hongwu Peng, Daniel Manu, Zhenglun Kong, Geng Yuan,
  Lei Yang, Shusen Wang, Hang Liu, and Caiwen Ding.
\newblock Hmc-tran: A tensor-core inspired hierarchical model compression for
  transformer-based dnns on gpu.
\newblock In {\em Proceedings of the 2021 on Great Lakes Symposium on VLSI},
  pages 169--174, 2021.

\bibitem{fang2021turbotransformers}
Jiarui Fang, Yang Yu, Chengduo Zhao, and Jie Zhou.
\newblock Turbotransformers: an efficient gpu serving system for transformer
  models.
\newblock In {\em Proceedings of the 26th ACM SIGPLAN Symposium on Principles
  and Practice of Parallel Programming}, pages 389--402, 2021.

\bibitem{peng2020selective}
Hongwu Peng, Balaji Narayanasamy, Asif~Imran Emon, Zhao Yuan, Rongxuan Zhang,
  and Fang Luo.
\newblock Selective digital active emi filtering using resonant controller.
\newblock In {\em 2020 IEEE International Symposium on Electromagnetic
  Compatibility \& Signal/Power Integrity (EMCSI)}, pages 632--639. IEEE, 2020.

\bibitem{wang2020benchmarking}
Zeke Wang, Hongjing Huang, Jie Zhang, and Gustavo Alonso.
\newblock {Benchmarking High Bandwidth Memory on FPGAs}.
\newblock {\em arXiv preprint arXiv:2005.04324}, 2020.

\bibitem{qi2021accommodating}
Panjie Qi, Yuhong Song, Hongwu Peng, Shaoyi Huang, Qingfeng Zhuge, and Edwin
  Hsing-Mean Sha.
\newblock Accommodating transformer onto fpga: Coupling the balanced model
  compression and fpga-implementation optimization.
\newblock In {\em Proceedings of the 2021 on Great Lakes Symposium on VLSI},
  pages 163--168, 2021.

\bibitem{peng2021binary}
Hongwu Peng, Shanglin Zhou, Scott Weitze, Jiaxin Li, Sahidul Islam, Tong Geng,
  Ang Li, Wei Zhang, Minghu Song, Mimi Xie, et~al.
\newblock Binary complex neural network acceleration on fpga.
\newblock In {\em 2021 IEEE 32nd International Conference on
  Application-specific Systems, Architectures and Processors (ASAP)}, pages
  85--92. IEEE, 2021.

\bibitem{peng2021accelerating}
Hongwu Peng, Shaoyi Huang, Tong Geng, Ang Li, Weiwen Jiang, Hang Liu, Shusen
  Wang, and Caiwen Ding.
\newblock Accelerating transformer-based deep learning models on fpgas using
  column balanced block pruning.
\newblock In {\em 2021 22nd International Symposium on Quality Electronic
  Design (ISQED)}, pages 142--148. IEEE, 2021.

\bibitem{yuan2021improving}
Geng Yuan, Zhiheng Liao, Xiaolong Ma, Yuxuan Cai, Zhenglun Kong, Xuan Shen,
  Jingyan Fu, Zhengang Li, Chengming Zhang, Hongwu Peng, et~al.
\newblock Improving dnn fault tolerance using weight pruning and differential
  crossbar mapping for reram-based edge ai.
\newblock In {\em 2021 22nd International Symposium on Quality Electronic
  Design (ISQED)}, pages 135--141. IEEE, 2021.

\bibitem{huang2022automatic}
Shaoyi Huang, Ning Liu, Yueying Liang, Hongwu Peng, Hongjia Li, Dongkuan Xu,
  Mimi Xie, and Caiwen Ding.
\newblock An automatic and efficient bert pruning for edge ai systems.
\newblock In {\em 2022 23rd International Symposium on Quality Electronic
  Design (ISQED)}, pages 1--6. IEEE, 2022.

\bibitem{peng2020design}
Hongwu Peng, Balaji Narayanasamy, Asif~Imran Emon, Zhao Yuan, Mustafeez~Ul
  Hassan, and Fang Luo.
\newblock Design and implementation of selective active emi filter with digital
  resonant controller.
\newblock In {\em 2020 IEEE Energy Conversion Congress and Exposition (ECCE)},
  pages 5855--5861. IEEE, 2020.

\bibitem{peng2021optimizing}
Hongwu Peng et~al.
\newblock {Optimizing FPGA-based Accelerator Design for Large-Scale Molecular
  Similarity Search}.
\newblock In {\em ICCAD}, 2021.

\bibitem{devlin2018bert}
Jacob Devlin, Ming-Wei Chang, Kenton Lee, and Kristina Toutanova.
\newblock {BERT: Pre-training of Deep Bidirectional Transformers for Language
  Understanding}.
\newblock In {\em 2019 ACL}, pages 4171--4186, 2019.

\bibitem{sanh2019distilbert}
Victor Sanh, Lysandre Debut, Julien Chaumond, and Thomas Wolf.
\newblock {DistilBERT, a Distilled Version of BERT: Smaller, Faster, Cheaper
  and Lighter}.
\newblock {\em arXiv preprint arXiv:1910.01108}, 2019.

\bibitem{liu2019roberta}
Yinhan Liu, Myle Ott, Naman Goyal, Jingfei Du, Mandar Joshi, Danqi Chen, Omer
  Levy, Mike Lewis, Luke Zettlemoyer, and Veselin Stoyanov.
\newblock Roberta: A robustly optimized bert pretraining approach.
\newblock {\em arXiv preprint arXiv:1907.11692}, 2019.

\bibitem{rajpurkar2016squad}
Pranav Rajpurkar, Jian Zhang, Konstantin Lopyrev, and Percy Liang.
\newblock Squad: 100, 000+ questions for machine comprehension of text.
\newblock In {\em EMNLP}, 2016.

\bibitem{dagan2010recognizing}
Ido Dagan, Bill Dolan, Bernardo Magnini, and Dan Roth.
\newblock Recognizing textual entailment: Rational, evaluation and
  approaches--erratum.
\newblock {\em Natural Language Engineering}, 16(1):105--105, 2010.

\bibitem{dolan2005automatically}
Bill Dolan and Chris Brockett.
\newblock Automatically constructing a corpus of sentential paraphrases.
\newblock In {\em Third International Workshop on Paraphrasing}. AFNLP, 2005.

\bibitem{zhang2020ternarybert}
Wei Zhang, Lu~Hou, Yichun Yin, Lifeng Shang, Xiao Chen, Xin Jiang, and Qun Liu.
\newblock {TernaryBERT: Distillation-aware Ultra-low Bit BERT}.
\newblock In {\em 2020 EMNLP}, pages 509--521, 2020.

\bibitem{qi2021accelerating}
Panjie Qi, Edwin Hsing-Mean Sha, Qingfeng Zhuge, Hongwu Peng, Shaoyi Huang,
  Zhenglun Kong, Yuhong Song, and Bingbing Li.
\newblock Accelerating framework of transformer by hardware design and model
  compression co-optimization.
\newblock In {\em 2021 IEEE/ACM International Conference On Computer Aided
  Design (ICCAD)}, pages 1--9. IEEE, 2021.

\end{thebibliography}










\end{document}